\documentclass[letterpaper, 10 pt, conference]{ieeeconf}  

\IEEEoverridecommandlockouts                              

\usepackage{graphics} 
\usepackage{epsfig} 

\usepackage{tabularx}
\usepackage{textcomp}
\usepackage{xcolor}
\usepackage{amsmath,amssymb,amsfonts}
\usepackage{subcaption}

\usepackage{tikz}

\newcommand\copyrighttext{%
  \footnotesize \textcopyright \the\year{} IEEE. Personal use of this material is permitted. Permission from IEEE must be obtained for all other uses, including reprinting/republishing this material for advertising or promotional purposes, collecting new collected works for resale or redistribution to servers or lists, or reuse of any copyrighted component of this work in other works.}

\newcommand\copyrightnotice{%
\begin{tikzpicture}[remember picture,overlay]
\node[anchor=south,yshift=10pt] at (current page.south) {\fbox{\parbox{\dimexpr0.75\textwidth-\fboxsep-\fboxrule\relax}{\copyrighttext}}};
\end{tikzpicture}%
}

\usepackage{booktabs} 
\title{\LARGE \bf
An Examination of Offline-Trained Encoders in Vision-Based Deep Reinforcement Learning for Autonomous Driving
}

\author{Shawan Mohammed$^{1}$, Alp Argun$^{2}$, Nicolas Bonnotte$^{1}$, Gerd Ascheid$^{2}$  
	\thanks{$^{1}$Akkodis Germany Solutions GmbH, Research Division,
		Flugfeld-Allee 12, D-71063 Sindelfingen
        {\tt\small shawan.mohammed@akkodis.com}
		{\tt\small nicolas.bonnotte@akkodis.com}
	}%
    \thanks{$^{2}$Department of Electrical Engineering, Institute for Communication Technologies and Embedded Systems (ICE), RWTH Aachen University, Aachen, Germany
		{\tt\small alp.argun@rwth-aachen.de}
		{\tt\small gerd.ascheid@ice.rwth-aachen.de}
	}
}

\begin{document}

\maketitle
\copyrightnotice

\thispagestyle{empty}
\pagestyle{empty}

\begin{abstract}
Our research investigates the challenges Deep Reinforcement Learning (DRL) faces in complex, Partially Observable Markov Decision Processes (POMDP) such as autonomous driving (AD), and proposes a solution for vision-based navigation in these environments. Partial observability reduces RL performance significantly, and this can be mitigated by augmenting sensor information and data fusion to reflect a more Markovian environment. However, this necessitates an increasingly complex perception module, whose training via RL is complicated due to inherent limitations. As the neural network architecture becomes more complex, the reward function's effectiveness as an error signal diminishes since the only source of supervision is the reward, which is often noisy, sparse, and delayed. Task-irrelevant elements in images, such as the sky or certain objects, pose additional complexities.
Our research adopts an offline-trained encoder to leverage large video datasets through self-supervised learning to learn generalizable representations. 
Then, we train a head network on top of these representations through DRL to learn to control an ego vehicle in the CARLA AD simulator. 
This study presents a broad investigation of the impact of different learning schemes for offline-training of encoders on the performance of DRL agents in challenging AD tasks. 
Furthermore, we show that the features learned by watching BDD100K driving videos can be directly transferred to achieve lane following and collision avoidance in CARLA simulator, in a zero-shot learning fashion.
Finally, we explore the impact of various architectural decisions for the RL networks to utilize the transferred representations efficiently. 
Therefore, in this work, we introduce and validate an optimal way for obtaining suitable representations of the environment, and transferring them to RL networks. 
We experimentally demonstrate the correlation between RL performance in the CARLA simulator and the quality of representations received by the RL agent. These results highlight the crucial role that feature extraction and state representation play in RL.
\end{abstract}


\section{Introduction}
In the domain of artificial intelligence, vision-based methods have arisen as a pivotal area of exploration. Leveraging camera images as primary inputs offers a versatile and cost-effective way to obtain a nuanced understanding of an environment by capturing colors, dimensions, and spatial orientation of objects. The recent advancements in deep learning and computer vision accentuate the capabilities of cameras, suggesting a potential reduction in reliance on expensive sensors like LiDAR.
However, the journey to achieve autonomy still faces numerous challenges. Autonomous driving operates primarily within the boundaries of Partially Observable Markov Decision Processes (POMDPs), where only a part of the environmental state is observed, called belief state. In an attempt to reduce the uncertainty in the belief state, historical information, sensor augmentation and data fusion methods are utilized. 

While it has been shown that this uncertainty is reduced when more information is added, it also poses several new problems.
The use of diverse sensors and their associated data fusion techniques amplify the complexity of the perception module. Hence, cutting-edge learning algorithms, datasets, and hardware solutions tailored for deep neural networks (DNN) and high-dimensional data streams are required.
This progression of complexity inevitably raises critical questions about the applicability of Deep Reinforcement Learning (DRL) in real-world AD scenarios.
End-to-end RL methods suffer from low sampling efficiency \cite{yarats} , which is especially critical for AD. High-resolution sensors produce vast amounts of data every second \cite{Baidu2017}, necessitating semantic processing at various layers \cite{Semantic}. The data often has spatiotemporal correlations that are essential to capture \cite{NatureSpatio}. 

\begin{figure}[!h]
	\begin{center}
		\includegraphics[width=1\linewidth]{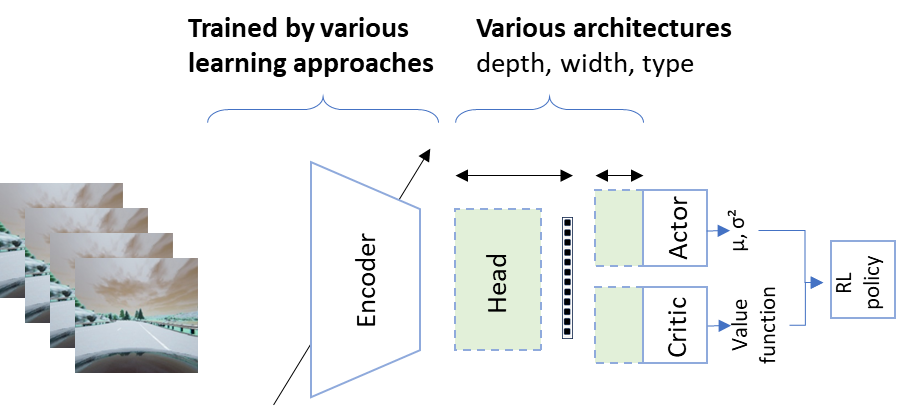} 
		\caption{Our framework with a frozen video encoder trained offline through various learning schemes and an ablation study with various head network architectures trained through DRL.}
		\label{fig:approach}
	\end{center}
\end{figure}

In this work, we address the challenges arising from the limitations of reward-driven representation learning of spatiotemporal input data for AD using DRL. We investigate the effects of various learning schemes for the offline-training of encoders for spatiotemporal representation learning, and the performance of RL agents with the transferred representations for AD, as depicted in Figure \ref{fig:approach}. Our experiments are conducted in the high-fidelity simulator CARLA \cite{Carla}, where RL agents control ego vehicles to achieve lane following and collision avoidance tasks. 

\newpage
\paragraph{Contribution} 
In this research, we investigate the significance of encoder representations for reinforcement learning for autonomous driving. Our first comprehensive ablation study aims to find the most suitable architectural design for the encoder's head, actor, and critic neural network. This careful exploration provides the foundation to ensure an optimal representation transfer, which is crucial for enhancing the performance.
Following this, we conducted a broad investigation of representation learning algorithms suitable for the offline-training of the encoder. For this purpose, we compared a variety of self-supervised learning approaches to learn from videos.
In this comparison, we used a generative model: variational autoencoder (VAE) \cite{VAE2016}, three joint embedding architectures: Momentum Contrast (MoCo) \cite{Moco2020} and Dense Predictive Coding (DPC) \cite{DPC2019}, where both utilize a contrastive loss, and Bootstrap Your Own Latent (BYOL) \cite{Byol2020} with a non-contrastive loss.
As an additional comparison, we incorporated YOLOPv2 \cite{yolopv2} for panoptic driving perception to compare representations trained for multiple perception tasks in a supervised manner with self-supervised learning representations on DRL performance.
We show that self-supervised representations learned by watching BDD100K driving videos can be transferred to CARLA driving simulator with no fine-tuning required.
Ultimately, this research points out how large video datasets can be leveraged to obtain encoded representations suitable learning to control an ego vehicle through DRL in a high-fidelity driving simulator.

\section{Related Work}

To improve the sample-efficiency of RL and learn latent dynamics, several works employ variational autoencoders (VAE) \cite{VAE-related} \cite{VAE-related2} applied to Google DeepMind infrastructure for physics-based simulation (DMControl) \cite{dmControl}.
Additionally, to tackle challenges in reward-driven, vision-based RL, several methodologies have adopted offline-trained encoders \cite{DRLOffline1}, \cite{DRLOffline2}, which use contrastive learning for training the encoders, and evaluated in DMControl tasks.
Yuan et al. \cite{pretrainedImageEncoder} employ earlier layers of a pre-trained image encoder trained on ImageNet \cite{imagenet} through supervision, and 
DMControl and CARLA. However, in this work we compare a variety of video encoders trained through contrastive, non-contrastive and generative models without any labels, and evaluate their performances on CARLA.
Additionally, Feichtenhofer et al. \cite{SSL3DRL} compare MoCo and BYOL for spatiotemporal representation learning for downstream computer vision tasks, where we compare them for control tasks in CARLA, and additionally employ a variational autoencoder as a generative model and another contrastive model, DPC \cite{DPC2019} for a more diverse investigation.

We conduct our study in the  the hyper-realistic environments presented by CARLA driving simulator, aiming to address real-world autonomous driving tasks. 
While earlier studies utilized vision-based RL methods in challenging environments, their visual data often lack the depth and complexity of real-world conditions. Real-world scenarios introduce nuances such as fluctuating weather, unpredictable pedestrian behavior, dynamic traffic, and occlusions, all with partial observability. 
The works of Toromanoff et al. \cite{CarlaSolved} and Carton et al. \cite{CarlaSolved2} leverage monocular cameras within CARLA, similar to us, to solve autonomous driving tasks by using imitation learning and highly informative reward functions. However, in this work, we do not use imitation learning and use a highly sparse and challenging reward function that offers less granular information to allow our agent to adapt to difficult situations, and generalize better. Additionally, we assess the performance implications of different learning schemes to offline-train encoders and architectural design decisions.

\section{Methodology}

Our methodology can be splitted into three major components: offline-trained encoder, head network and DRL model.
First we train an encoder to learn generalizable features from an offline video dataset to be transferred to our DRL model. This encoder is trained offline based on various state of the art algorithms from computer vision.
Then, the representation of the encoder is used for decoupled DRL to learn to control a vehicle in a driving simulator.
Finally, we perform an ablation study for designing the head network to transfer the learned representations optimally for vision-based navigation.  

\subsection{1. Offline-trained Encoder}

In order to utilize large video datasets and transfer this knowledge to our DRL network, we employ an encoder so that our DRL network can learn the ego vehicle control from the predicted features, instead of from pixels.

We use a sequence of four images as input to our agent. 
Hence, for our encoder, we use a 3D-ResNet18 \cite{resnet}, due to its feature extraction capabilities. This architecture provides an optimal balance between speed and performance for our study. Thus, while preserving adequate complexity, its size allows adequate training and inference time.
We trained our encoders offline on the the Berkeley Deep Drive (BDD 100K) dataset \cite{BDD2018} with no annotations, in a self-supervised manner. 
All our encoders use an input of size $4x3x128x128$, due to four stacked RGB frames of 128 pixels. 
This training process seeks to test the encoders' generalizability: \textit{Can they recognize patterns valid beyond their initial dataset?}
Our research aims to determine which learning scheme, guided by its inherent loss function, can provide the best representations to learn the control task in CARLA simulator.
The Results chapter provides a detailed evaluation of each encoder, presenting an empirical answer to our central question.
For a detailed list of hyperparameters and a description of the losses, we refer to our GitHub repository \cite{SSL3DRL}.

\paragraph{Momentum Contrast (MoCo)} 
MoCo \cite{Moco2020} utilizes the InfoNCE contrastive loss \cite{infoNCE}, which learns to pull together augmented views of the same sample, and push away the embeddings from other samples. Hence, MoCo objective maximizes the similarity $sim$ with the positive keys $\{k^+\}$ and minimizes the similarity with negative keys $\{k^-\}$ as in Equation \ref{e:3}.

\begin{samepage}
	\begin{equation}
		\label{e:3} 
		{L}_{q} = -log\frac{{\sum}_{k \in \{ k^+ \}}exp(sim(q,k)/\alpha)}{{\sum}_{k \in \{ k^+ , k^-\}}exp(sim(q,k)/\alpha)}	
	\end{equation}
	\begin{equation*}
		\text{with : } \text{ } \text{ } \text{ } \text{ } \text{ } \text{ } sim(q^T,k) = \frac{q^T k}{\|q\| \|k\|}	
	\end{equation*}
	\begin{tabularx}{\linewidth}{@{}XX@{}}
		\begin{itemize}
			\item $q$  is the encoded query sample
			\item $\alpha$ is a temperature coefficient, which we set 0.1 
		\end{itemize}
		&
		\begin{itemize}
			\item $k$ are encoded samples, $\{k_0 , k_1 , k_2 , ...\}$, as keys of the dictionary
		\end{itemize}
	\end{tabularx}
\end{samepage}

\paragraph{Bootstrap Your Own Latent (BYOL)} BYOL utilizes a non-contrastive loss, which aims for the consistency in representation across different augmentations of the same sample \cite{Byol2020}. 
Without a contrastive objective, in order to prevent representation collapse, where the model predicts the same constant vector regardless of the input \cite{collapse}, BYOL implements an extra multi-layer perceptron predictor head on top of the query encoder.
BYOL uses a similarity loss, a simple mean square error (MSE) between the representations of the query encoder $q_{\theta}(z_{\theta})$ and the key encoder $z'_{\xi}$, as shown in Equation \ref{e:5}.

\begin{samepage}
	\begin{equation}
		\label{e:5} 
		L_{\theta, \xi} \overset{\Delta}{=} \| \overline{q_{\theta}}(z_{\theta}) -  \overline{z'_{\xi}}  {\|}_{2}^{2}
	\end{equation}
\end{samepage}

To apply MoCo and BYOL for video, we used their adaptations to spatiotemporal representation learning, through 3D convolutions and an additional temporal persistency objective \cite{spatiotemporal2021}, so that the clips from the same video have similar embeddings, and the clips from different videos have distant embeddings. Furthermore, MoCo and BYOL utilize a momentum encoder, whose weights are a moving averages of the original encoder.

\paragraph{Dense Predictive Coding (DPC)} 
DPC learns representations by recurrently predicting future representations \cite{DPC2019}. 
Similar to MoCo, DPC is a contrastive learning approach, utilizing the InfoNCE loss (Eq. \ref{e:3}).
However, DPC differs from MoCo by not utilizing a momentum encoder, but aggregating the predicted representations over the temporal axis using recurrent neural networks.  
Hence, it implicitly enforces a temporal coherence to learn the inherent temporal dynamics of the data. 
			

\paragraph{Variational Autoencoder (VAE)} 
A VAE consists of two main components: an encoder, which maps the input data to a lower-dimensional latent space, and a decoder, which reconstructs the input data from the latent representation \cite{VAE2016}.
While the reconstruction loss encourages compact and efficient encoding of the input data, the Kullback-Leibler (KL) divergence term pushes the learned latent variables to adhere to a known distribution, enabling smoother interpolation and generalization in the latent space, as in Equation \ref{e:41}. To make the autoencoder comparable to the other approaches used in this work, we adapted it to take a sequence of four input images. 
In contrary to MoCo, BYOL and DPC, which are joint embedding architectures, VAE is a generative model, which learns the representations with a reconstruction objective.  

\begin{samepage}
	\begin{equation}
		\label{e:41} 
		L_{\text{VAE}} = \| x - \hat{x} \|_2^2 + D_{\text{KL}}(q(\mathbf{z}|x) \| p(\mathbf{z})) 
	\end{equation}
	\begin{tabularx}{\linewidth}{@{}XX@{}}
		\begin{itemize}
			\item Original data \( x \) and its reconstruction \( \hat{x} \).
		\end{itemize}
		&
		\begin{itemize}
			\item Approximate posterior \( q(\mathbf{z}|x) \) and the prior \( p(\mathbf{z}) \).
		\end{itemize}
	\end{tabularx}
\end{samepage}

\paragraph{YOLOPv2} 
YOLOPv2 presents an efficient network that performs object detection, driveable area segmentation, and lane line detection simultaneously \cite{yolopv2}. The network consists of three heads to achieve three different perception tasks and use the same backbone and neck as a shared encoder. The backbone is based on E-ELAN \cite{yoloEELAN}, which focuses on multiple tasks and employs an attention mechanism \cite{attentionNIPS}. Furthermore, YOLOPv2 on BDD100K is trained in a supervised manner for these tasks using labeled data. 

Despite the major differences in network architecture and the training scheme with our other encoders, we chose the encoder of YOLOPv2 as a benchmark for our scalable and efficient self-supervised learning-based encoders, based on their impact on DRL performance for AD. 


\subsection{2. Deep Reinforcement Learning for Autonomous Driving}
\begin{figure}[!h]
	\begin{center}
		\includegraphics[width=1\linewidth]{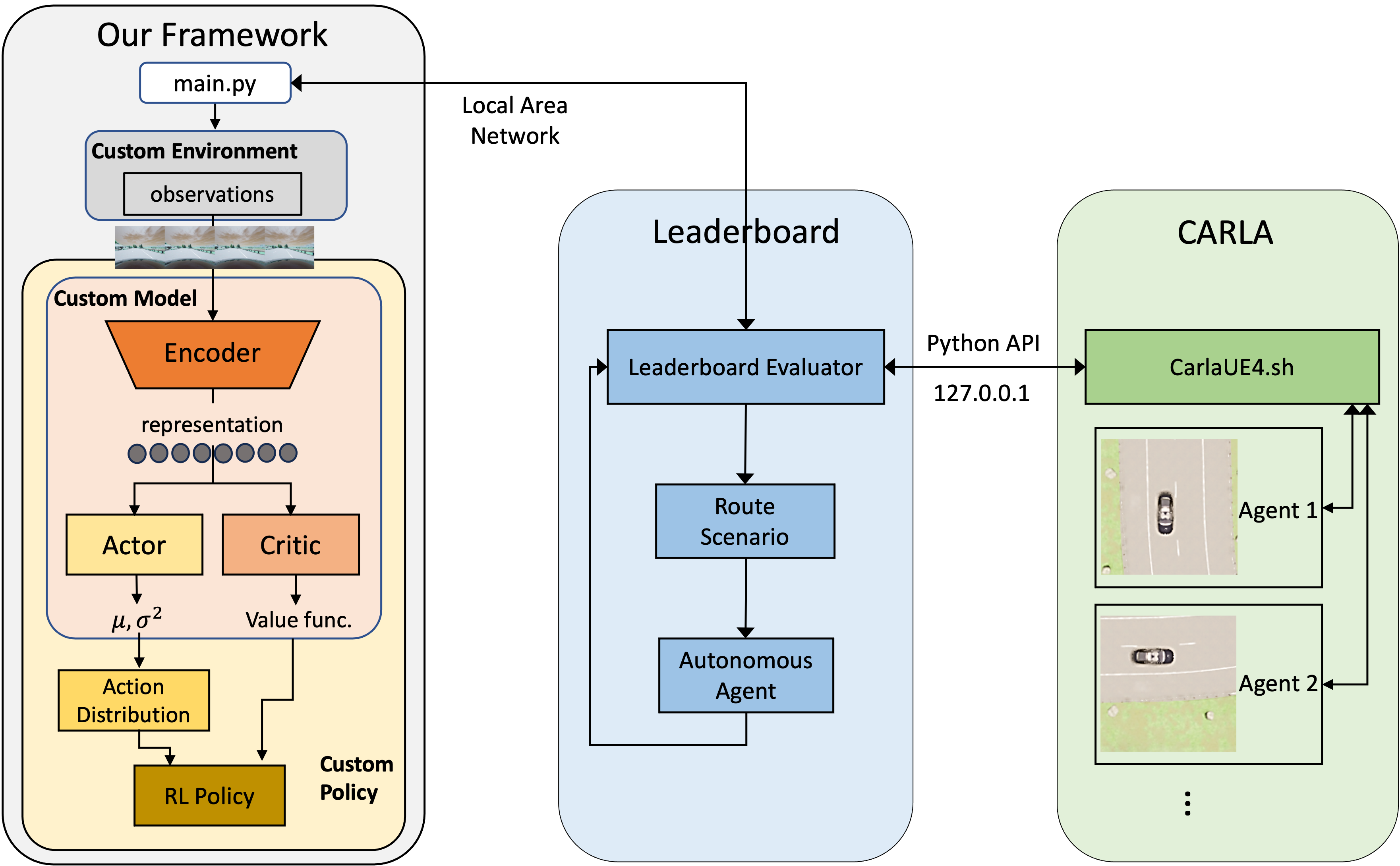} 
		\caption{Our framework for AD and its communication with CARLA and CARLA Leaderboard.}
		\label{fig:drl_plattform}
	\end{center}
\end{figure}

For this study, we adapted our existing DRL framework \cite{unifiedAD-shawan-alp} and used it for the validation of our encoders for vision-based navigation in CARLA driving simulator. Figure \ref{fig:drl_plattform} shows an overview of the DRL framework used with all its integral parts. The framework consists of three main modules: environment, leaderboard, and DRL framework. CARLA leaderboard standardizes traffic scenario definitions and provides a benchmark to compare the performance of different approaches. We integrate the offline trained encoders and implement an additional head network to predict vehicle controls into our DRL framework so that the RL agent can learn higher-level decision-making and control policies from visual abstractions rather than learning multiple perception tasks.  
By freezing the encoder layers, we prevent them from being trained by the RL loss function, resulting in a visual abstraction, i.e., representation learning based RL training configuration. This configuration trains only the actor and critic NNs with RL loss and uses representations as inputs $s_t$ from the frozen encoder module. We standardized the output of the encoders, so they predict representations of same dimensionality, regardless of the training scheme. We use proximal policy optimization for RL loss, as in Equation \ref{e:1} \cite{PPO2017}.

\begin{samepage}
	\begin{equation}
		\label{e:1} 
		{L}^{CLIP}(\theta) =\hat{E}[min(r_t(\theta)\hat{A}_t, clip(r_t(\theta), 1-\epsilon, 1+ \epsilon)\hat{A}_t)]	
	\end{equation}
	\begin{equation*}
		\text{with : } \text{ } \text{ } \text{ } \text{ } \text{ } \text{ } r_t(\theta) = \frac{\pi_{\theta}(\alpha_t | s_t)}{\pi_{\theta_{old}}(\alpha_t | s_t)}	
	\end{equation*}	
	\begin{tabularx}{\linewidth}{@{}XX@{}}
		\begin{itemize}
			\item $\theta$  is the policy parameter
			\item $\hat{E}$  denotes the empirical expectation over timesteps
			\item $\hat{A}$  is the estimated advantage at time $t$
			\item $\epsilon$  is a clipping parameter, we used 0.1
		\end{itemize}
		&
		\begin{itemize}
			\item $r_t$  is the ratio of the probability under the new $\pi_{\theta}$ and old policies $\pi_{\theta_{old}}$, respectively
			\item $\alpha$  is the action at time $t$
			\item $s_t$  is the input of the actor and critic at time $t$
		\end{itemize}
	\end{tabularx}
\end{samepage}

The actor and critic modules have the same architecture, differing only in the final layers. 
The output dimension of the actor module is $2x2$ to predict a mean ($\mu$) and a variance ($\sigma^2$) for both the steering and acceleration controls of the ego-vehicle, where both steering and acceleration are defined in a continuous action space.
The output dimension of the critic module is one and predicts only the value function. Table \ref{tab:heisetabelle} shows a detailed overview of the RL and NN hyperparameters as well as the hardware configurations used for our training and can also be looked up in our Github repository.

\begin{table*}[h]
	\centering
	\scriptsize
	\caption{Hyperparameter used for RL training.}
	\label{tab:heisetabelle}
	\begin{tabular}{lll}
		\toprule
		\multicolumn{2}{c}{RL trainings parameter part 1} \\
		\cmidrule(r){1-2}
		Parameter 			& Value \\
		\midrule
		RL algorithm	& PPO \\
		Sequence length 	& 4 \\
		KL coefficient	& 0.3 \\
		KL target coefficient	& 0.03 \\
		Nr. of sgd iterations & 3 \\
		$\epsilon$ clip param & 0.1 \\
		Vf clip 		& 10 \\
		Action normalization 	& True \\
		Sgd mini batch size & 128\\
		Nr. of sgd iterations & 3 \\
		Lambda 			& 0.95 \\
		Entropy coefficient & 0.0 \\
		\bottomrule
	\end{tabular}
	\begin{tabular}{lll}
		\toprule
		\multicolumn{2}{c}{RL trainings parameter part 2} \\
		\cmidrule(r){1-2}
		Parameter 			& Value \\
		\midrule
		Gamma 			& 0.9 \\
		Batch size 			& 640 \\
		Rollout fragment length & 64 \\
		Horizon				& 64 \\
		Optimizer 			& adam \\
		Learning rate 		& 2e-5 \\
		GAE used 		& True \\
		Initializer			& He-normal \\
		Reqularizer 		& L1 and L2 \\
		GPU 				& 2xRTX 3090TI \\
		CPU 				& i9-10900K \\
		RAM / HDD 				& 64GB / 2TB \\
		\bottomrule
	\end{tabular}
	\begin{tabular}{lll}
		\toprule
		\multicolumn{2}{c}{Equal NN architecture for actor and critic (AC)} \\
		\cmidrule(r){1-2}
		Parameter 			& Value \\
		\midrule
		AC input shape MoCo/BYOL & (4,4,2048) \\
		AC input shape YOLOPv2 & (4,4,2048) \\
		AC input shape (DPC) & (8,8,128) \\
		No. of conv.layer (CL) & 2 \\
		Kernel size CL1/CL2 & 2/2 \\
		Stride CL1/CL2 & 1/1 \\
		Channels CL1/CL2 & 1024/512 \\
		No. of FC layer (FCL) & 2 \\
		Units FC1/FC2 & 400/100 \\
		Activation function (all) & Relu \\
		Critic output layer (units) & 1 (linear)\\
		Actor output layer (units) & 2 (linear)\\
		\bottomrule
	\end{tabular}
\end{table*}


\paragraph{Environment setup}
We conducted experiments in CARLA focusing on lane following and collision avoidance. The evaluation was limited exclusively to the steering control of an ego vehicle, with acceleration control taken over by the autopilot. This approach was chosen because learning acceleration control requires additional information, such as variable speed limits on different road types. In addition, successful acceleration control would place increased demands on NN, particularly regarding data fusion within the networks and adaptation of input data. These additional requirements could change the nature of our ablation study and complicate the investigation of the impacts of different offline encoders. Therefore, the focus on steering control was chosen in this study. 

\begin{figure}[!h]
	\begin{center}
		\includegraphics[width=0.7\linewidth]{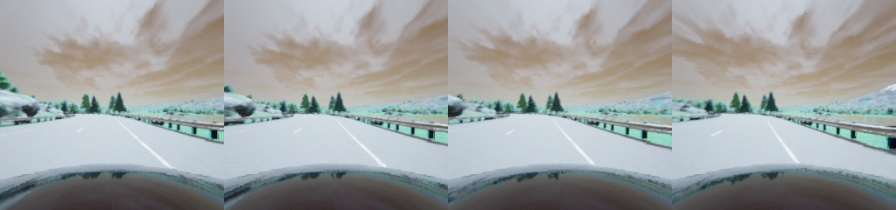}
		\caption{Observation space with a sequence of four frames.}
		\label{fig:sequence_iamges}
	\end{center}
\end{figure}

The observation space for all encoders is uniform, with a frame size of $3x128x128$ and four stacked RGB frames. An example of the observation space used is shown in Figure \ref{fig:sequence_iamges}. Due to resource constraints, we ran the CARLA experiments with four agents. We initialized the agents on training routes 31, 35, and 46 provided by the CARLA Leaderboard \cite{Leaderboard}. The roads can be curvy or straight depending on random spawn location. Here, an episode is always exactly 400 steps long unless there is a collision.

\paragraph{Reward function} The reward function for RL training consists of three components. The first component, $r_l$, relates to lane violation penalties, and the second component, $r_c$, relates to collision penalties. Both components are set to 1 when the ego vehicle crosses a lane or collides with another vehicle, walls, or pedestrians. In all other cases, it is 0, as shown in Eq. \ref{e:2}. Since all reward components are negative, the maximum achievable reward is zero. Each component has its own coefficients: $c_l$ = 2 and $c_c$ = 30, which are used to distinguish the priority of the events.  

\begin{equation}
	\label{e:2} 
	r_{reward} = - c_{l}r_{l} - c_{c}r_{c} 	
\end{equation}

\noindent
\begin{tabularx}{\linewidth}{@{}XX@{}}
	\begin{equation*}
		r_{l} = 
		\begin{cases}
			0, & \text{if } $kept lane$\\
			1, & \text{if } $left lane$
		\end{cases} 
	\end{equation*}
	&
	\begin{equation*}
		r_{c} = 
		\begin{cases}
			0, & \text{if } $no collision$\\
			1, & \text{if } $collision$
		\end{cases} 
	\end{equation*}
\end{tabularx}

The reward function, $r_{reward}$, is intentionally formulated very simply compared to other works solving the same task \cite{CarlaSolved}, \cite{CarlaSolved2}. In this way, we challenge the RL optimization for learning complex policies by a simple and sparse but goal-oriented reward signal, as we saw that this allows a more natural drive in our previous work \cite{unifiedAD-shawan-alp}.

\paragraph{Implementation Details}

Our framework consists of a distributed RL training on multiple instances of CARLA simulator to collect training samples more efficiently.
In order to standardize the models we compared, we converted all models to the Open Neural Network Exchange (ONNX) standard as it provides a unified representation for machine learning models \cite{ONNX1}, regardless of the models were implemented in PyTorch or TensorFlow.


\subsection{3. Ablation Study on Designing a Head Network for Vision-based Navigation}
When adapting encoders, initially trained on SSL objectives, to downstream vision-based navigation tasks through DRL, it's pivotal to transfer the representations to a task-specific head network. To design the head network for our encoder, we perform an ablation study examining three different architectures for the head network.

\begin{figure}[!h]
	\begin{center}
		\includegraphics[width=1\linewidth]{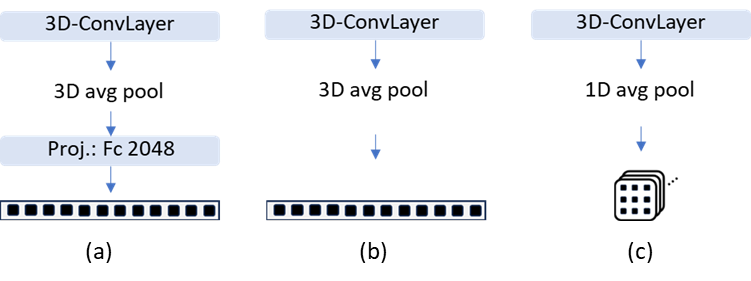}
		\caption{Example inputs consisting of an image sequence.}
		\label{fig:head_iamges}
	\end{center}
\end{figure}

\paragraph{Direct Feature Extraction} In this setup, the 2048 features from the projection layer are flattened and used directly without the prediction layer, as in Figure \ref{fig:head_iamges}(a). By omitting the last prediction layer of the original head, which is highly adapted to the original self-supervised learning task, more general features can be transferred. This 1D vector is passed to the actor and critic networks, which consist entirely of fully connected (FC) layers. 

\paragraph{Conv-Layer with 3D Averaging} Alternatively, we obtained the feature maps directly from the last convolutional layer and applied a 3D averaging layer. This 3D averaging layer was central in this stage, proficiently shaping these feature maps into a compressed 1D feature vector of 2048 features, as in Fig. \ref{fig:head_iamges}(b). Similar to the "Direct Feature Extraction" method, the compactness of this vector enabled our actor and critic networks to utilize FC layers exclusively.

\paragraph{Temporal Axis Reduction} In this strategy, rather than extensive compression, we employ a 1D averaging layer along the temporal axis to pool the features from multiple frames. Consequently, this results in feature maps of dimensions: channels, height, and width, as in Fig. \ref{fig:head_iamges}(c). Given the spatial nature of these representations, both actor and critic networks are then equipped with convolutional layers to process received spatial features efficiently.

\begin{figure*}
	\makebox[\textwidth][c]{%
		\begin{subfigure}[b]{.21\paperwidth}
			\centering
			\includegraphics[width=.95\textwidth]{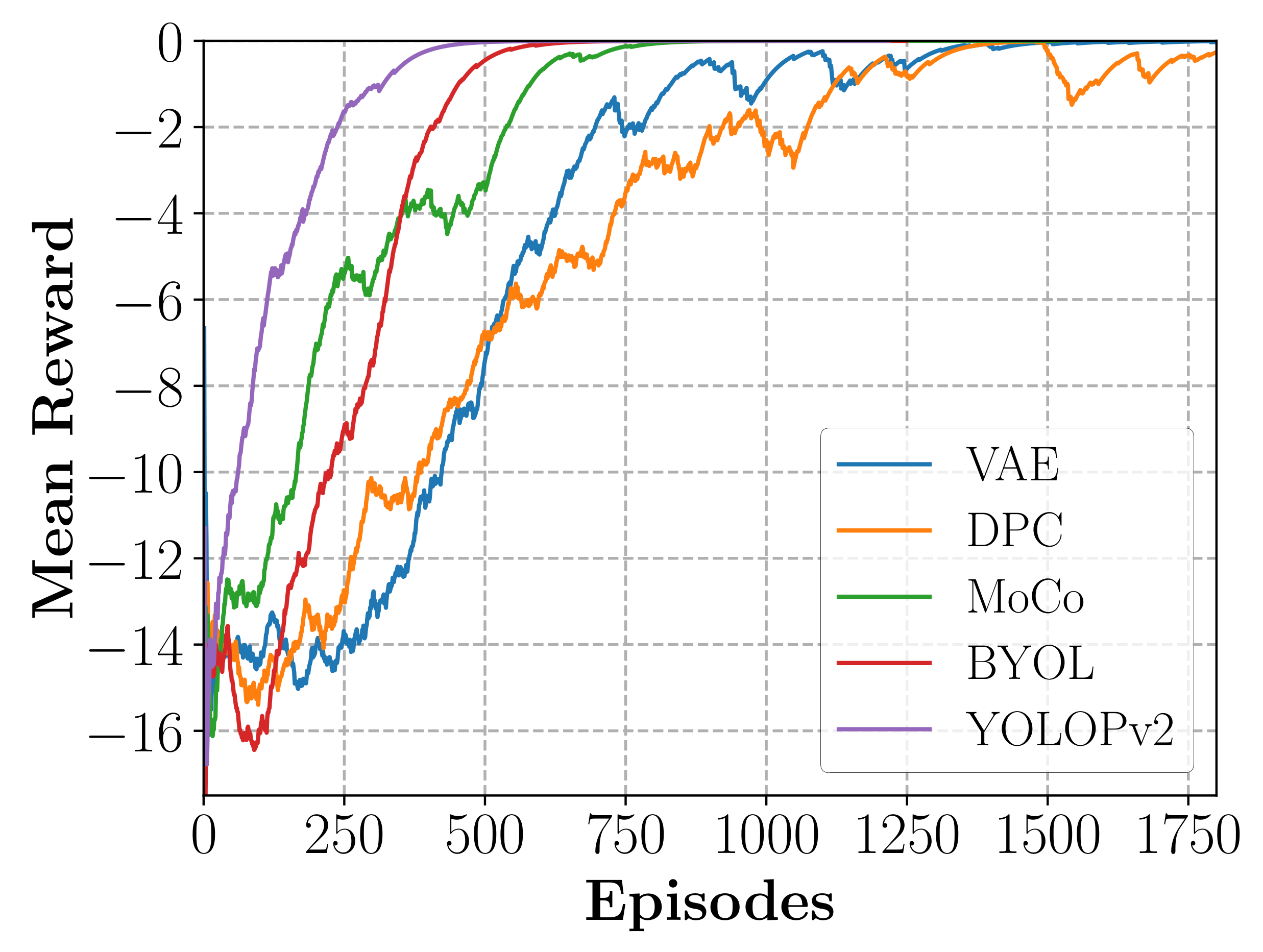}
			\caption{Mean reward/episode}
			\label{fig:vanillaDecoup}
		\end{subfigure}%
		\begin{subfigure}[b]{.205\paperwidth}
			\centering
			\includegraphics[width=.95\textwidth]{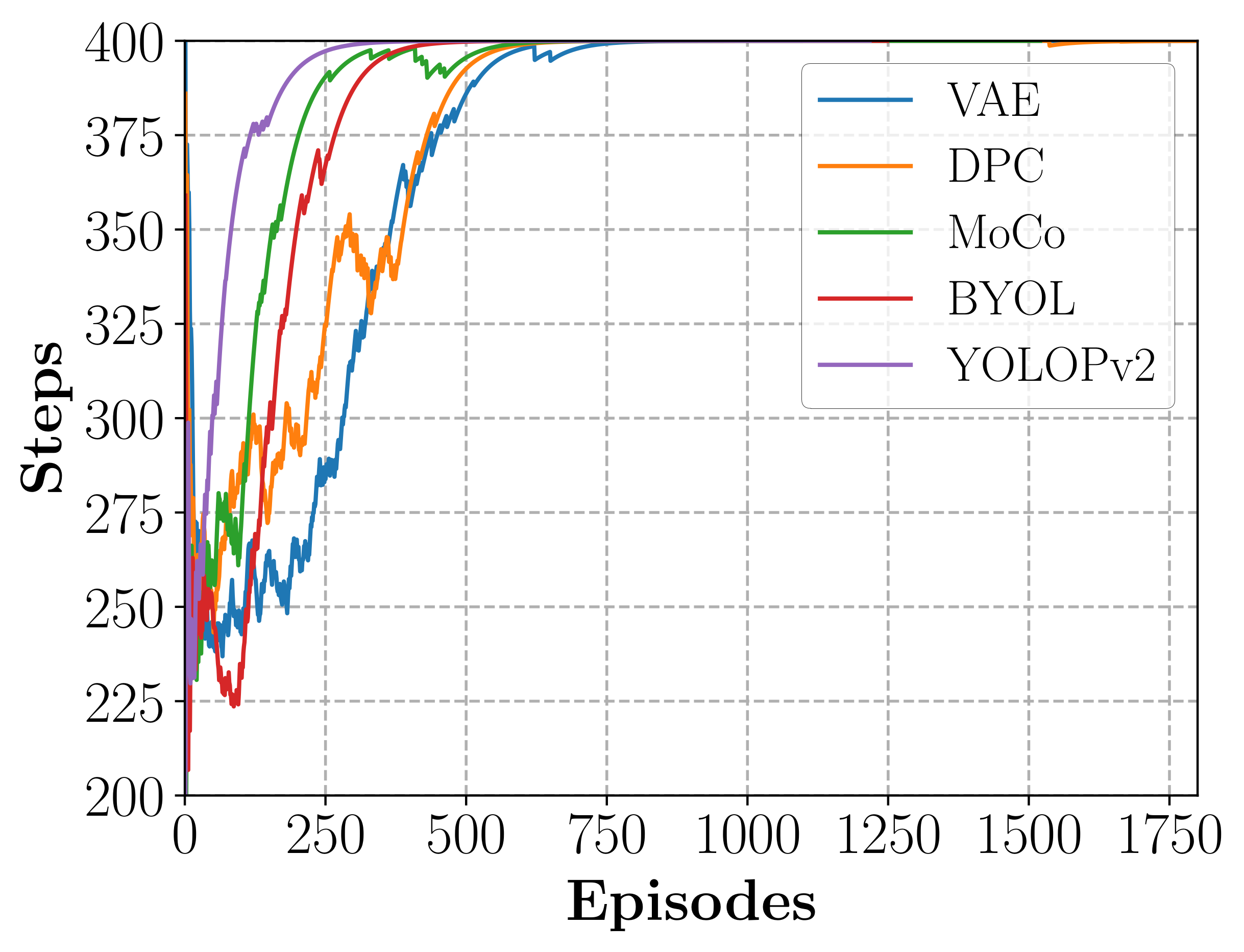}
			\caption{No. of steps/episode}
			\label{fig:vanillaDecoup2action}
		\end{subfigure}%
		\begin{subfigure}[b]{.21\paperwidth}
			\centering
			\includegraphics[width=.95\textwidth]{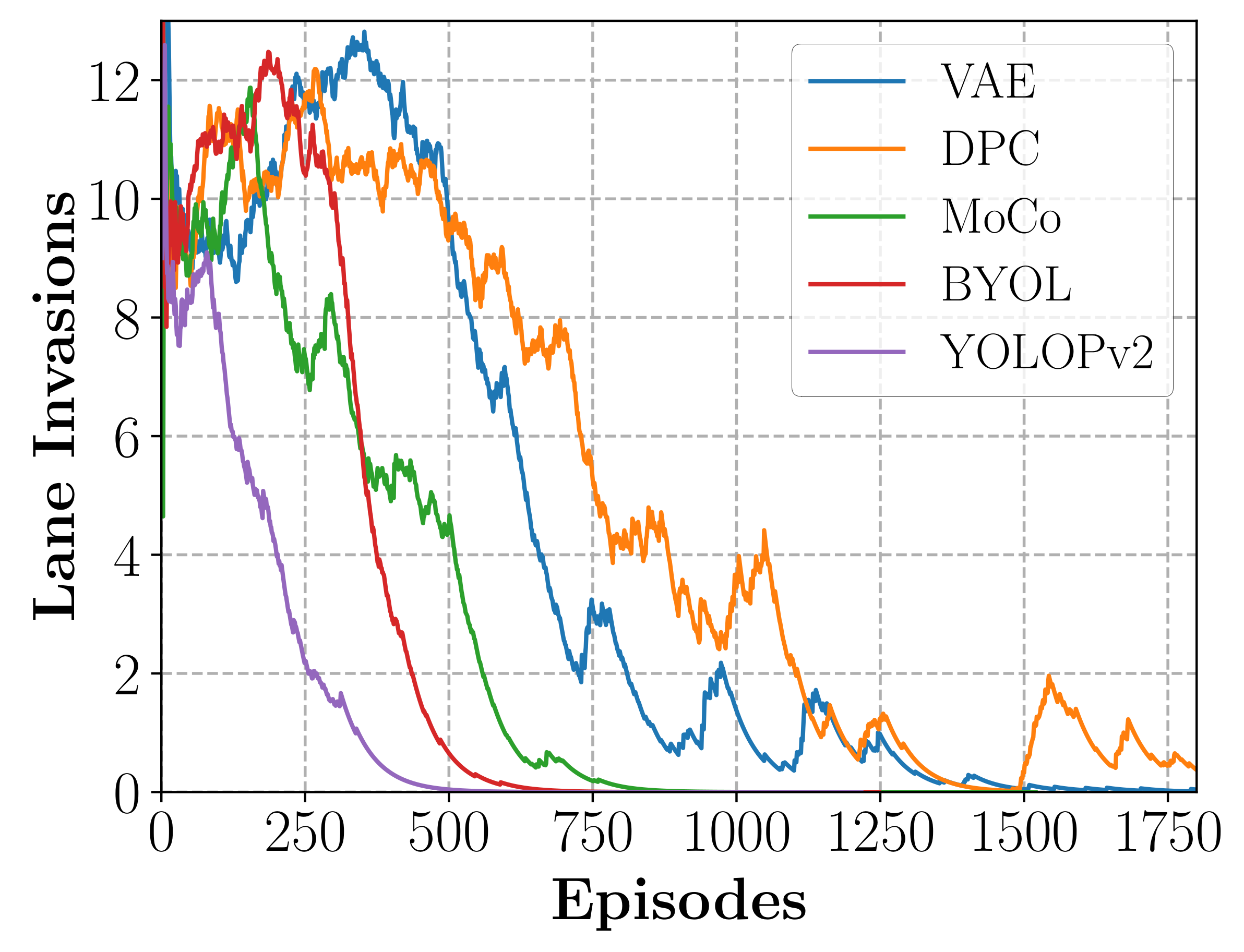}
			\caption{No. of lane invasions/episode}
			\label{fig:2ActionModi}
		\end{subfigure}%
	}
	\caption{Evaluation and comparison of different trained encoders}
\end{figure*}

\begin{figure*}
	\makebox[\textwidth][c]{%
		\begin{subfigure}[b]{.21\paperwidth}
			\centering
			\includegraphics[width=.95\textwidth]{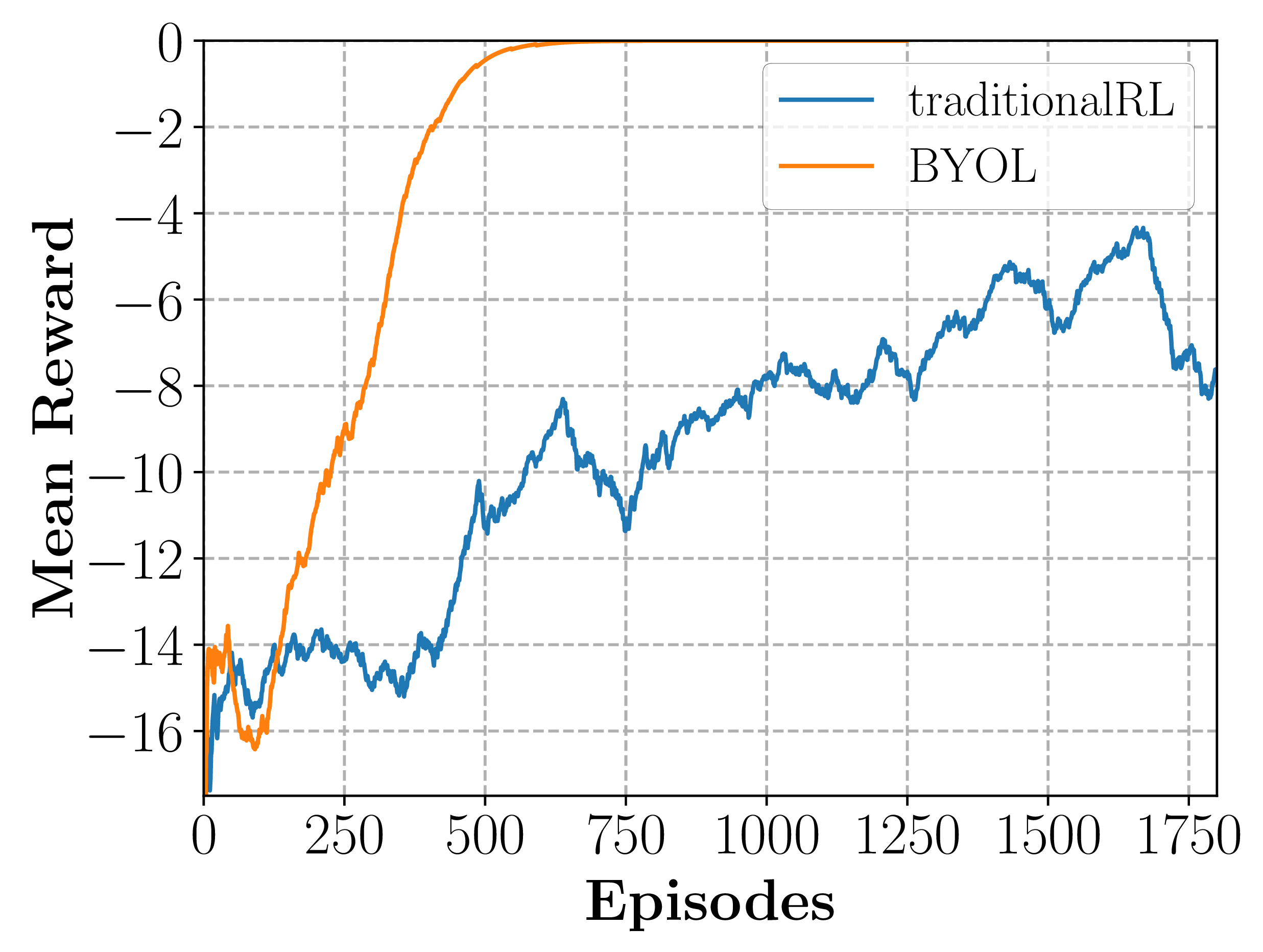}
			\caption{Mean reward/episode}
			\label{fig:whyDecoupledRL1}
		\end{subfigure}%
		\begin{subfigure}[b]{.205\paperwidth}
			\centering
			\includegraphics[width=.95\textwidth]{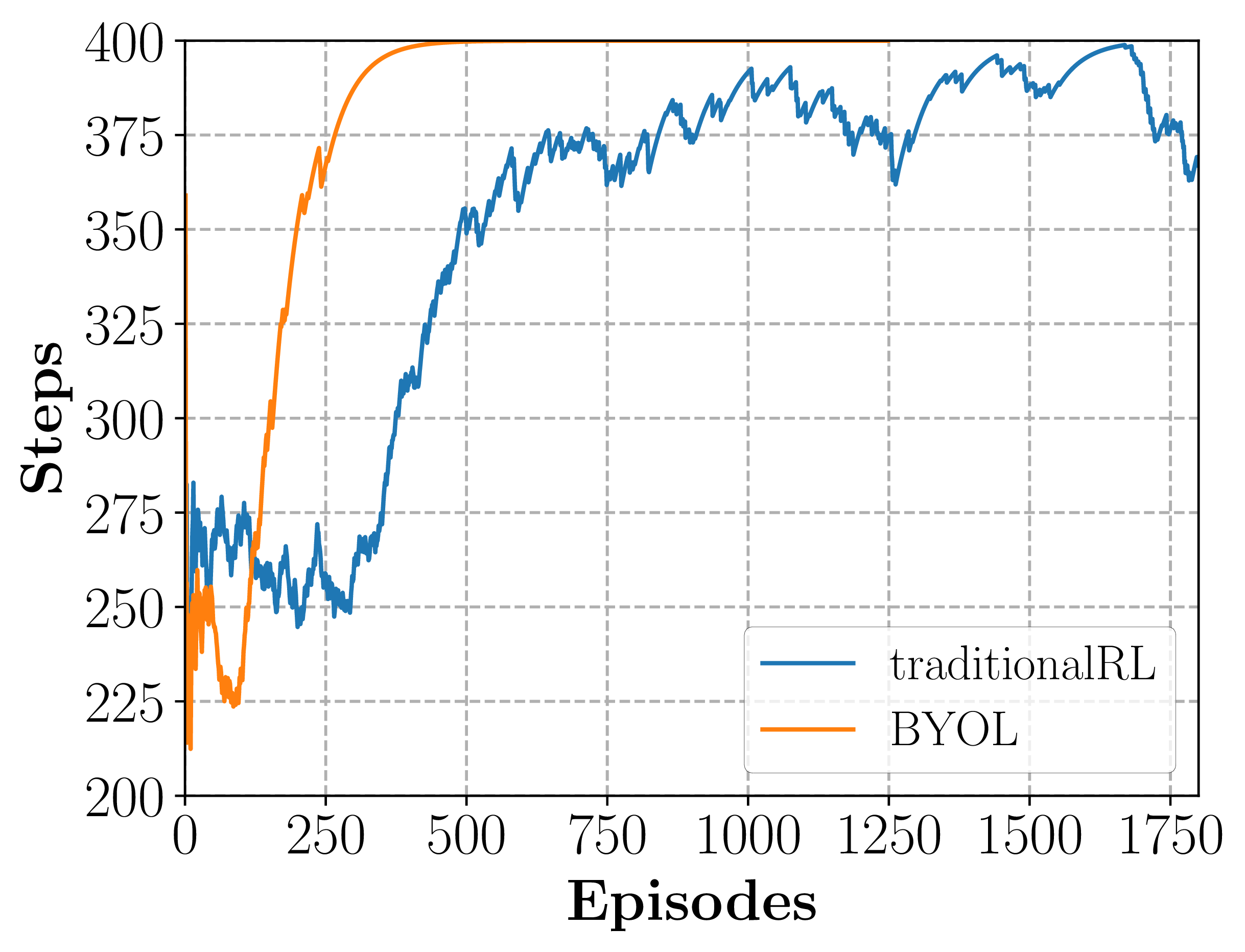}
			\caption{No. of steps/episode}
			\label{fig:whyDecoupledRL2}
		\end{subfigure}%
		\begin{subfigure}[b]{.21\paperwidth}
			\centering
			\includegraphics[width=.95\textwidth]{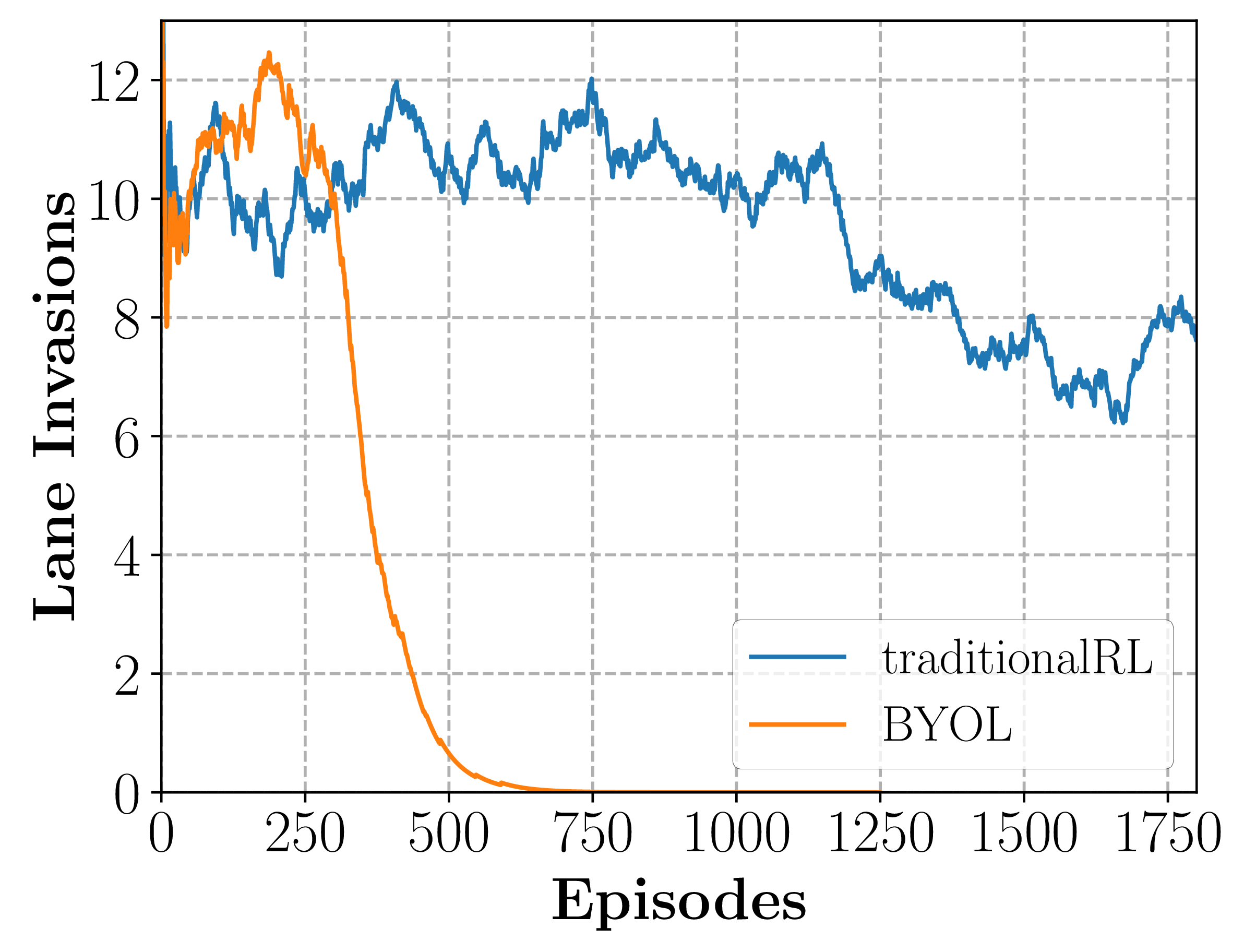}
			\caption{No. of lane invasions/episode}
			\label{fig:whyDecoupledRL3}
		\end{subfigure}%
	}
	\caption{Comparison of end-to-end RL versus training only the head network of a BYOL encoder.}
\end{figure*}

\section{Results}
In this chapter, we first present outcomes from our architectural design search for the encoder head, actor, and critic. Using the optimal configuration, we then evaluate the performance of the specified encoders. Our best-trained encoder's performance, utilizing the top architectural design, is subsequently compared against an end-to-end DRL approach.
Our results are showcased in two formats:
\begin{itemize}
	\item Cumulative reward per episode, adhering to RL conventions.
	\item Metrics from the CARLA leaderboard, indicating lane invasions and steps per episode.
\end{itemize}
Each experiment was run three times; we depict the median, smoothed with a 100-step moving average, for clarity.

\subsection{Architectural Design Search}
\begin{figure}[!h]
	\begin{center}
		\includegraphics[width=0.9\linewidth]{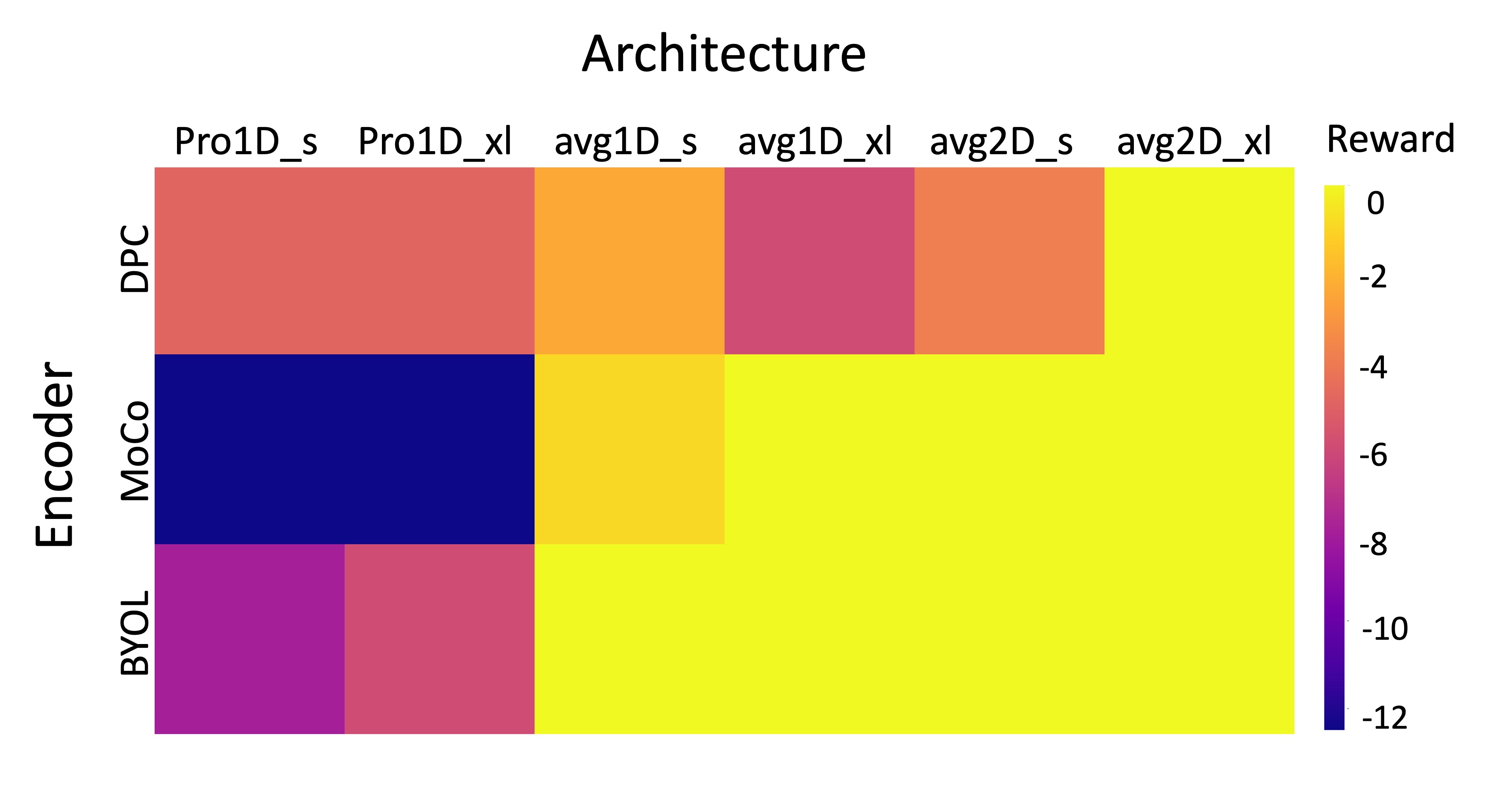}
		\caption{Heatmap visualization of encoder performance}
		\label{fig:heatmap}
	\end{center}
\end{figure}

The results of the architectural design search are visualized as a heatmap in Fig. \ref{fig:heatmap}. Each of the head networks on top of the frozen encoders was trained for 1800 episodes. After normalization, we plotted the highest achieved reward. In the subsequent descriptions, the two variations of each of the three methods are labeled with a suffix: $s$ indicates the smaller neural network, while $xl$ designates the larger one.
Both variants of the "Direct Feature Extraction" method, labeled as $Pro1D\textunderscore s$ for the smaller version and $Pro1D\textunderscore xl$ for the larger network, perform poorly across all three encoder types examined. The "Temporal Axis Reduction" variants ($avg2D\textunderscore s$ and $avg2D\textunderscore xl$) exhibit the best performance, with DPC performing well only when paired with the larger network. The "Conv-Layer with 3D Average" versions ($avg1D\textunderscore s$ and $avg1D\textunderscore xl$) perform variably depending on the encoder, obtaining mediocre to high rewards.

\subsection{Analysis of Different Learning Schemes to Train Encoders}
Figure \ref{fig:vanillaDecoup}, \ref{fig:vanillaDecoup2action}, and \ref{fig:2ActionModi} showcase evaluations for understanding how different trained encoders influence RL performance. Notably, most RL trainings with offline-trained encoders succeeded in the task, with DPC being an exception. The DRL training of the head network with DPC as the encoder failed to converge, consistently committing lane invasions even beyond 1800 episodes. A successful driving behavior requires zero lane invasions and zero collisions to achieve the maximum possible reward, which is zero. The VAE encoder exhibited this, albeit with better convergence than DPC. MoCo and BYOL succeeded in the task, demonstrating a consistent driving behavior. Their primary distinction was in convergence speed: BYOL reached zero rewards in just 700 episodes, while MoCo took 850 episodes. Due to its supervised and multi-task learning scheme, and more complex E-ELAN backbone, YOLOPv2 encoder showed a faster convergence and reached a reward of zero in merely 500 episodes, as expected. However, BYOL is a close second, which shows the strength of BYOL representations, even though trained without labels and with a smaller backbone, 3D-ResNet18.
Figure \ref{fig:whyDecoupledRL1} highlights the significant performance difference between training with frozen and offline trained encoders, and traditional end-to-end RL. Despite using similar architectures and configurations—based on the same DRL framework and hyperparameters from Table \ref{tab:heisetabelle}—differences are evident. End-to-end RL fails at the lane following and collision avoidance tasks, and thus, can only reach a reward of -4. In contrast, our BYOL encoder with a trained head network not only learns an effective policy but also attains zero rewards after approximately 700 episodes, which indicates of a flawless driving behavior without any lane violations or collisions, as shown by Figures \ref{fig:whyDecoupledRL2} and \ref{fig:whyDecoupledRL3}.

\section{Conclusion}

In this work, we examine multiple self-supervised spatiotemporal representation learning approaches to train an encoder for AD using DRL. We train our encoders on the BDD100K driving dataset, and design a head network on top of these encoders to predict vehicle controls in CARLA driving simulator. Then, we train the head network through DRL to learn the lane following and collision avoidance tasks.
Our experiment results showed that learning by reconstruction produces weaker features for perception as also shown in \cite{reconstructionPerception}. On the other hand, joint embedding architectures learned stronger features, which are also useful for control tasks.
As our non-contrastive approach, BYOL was the best performing SSL approach, we believe contrastive learning can be harmful to learned representations, as some videos in the dataset can be still similar to each other, hence, include similar features. However, contrastive learning pushes these embeddings away from each other.
Furthermore, even though Transformer-based approaches proved their strength in vision tasks \cite{MAE-he}
we show that a relatively lightweight ConvNet, such as a 3D-ResNet18 can capture useful features from a sequence of frames that can be useful for control.
Our investigation for designing the head network showed that when the representation quality is high such as the encoder trained by BYOL, the impact of the design of the head network is less significant. However, with lower representation quality, such as the encoder trained by DPC, the design of the head network is important for succeeding in the control task.
Finally, self-supervised learning can learn generalizable features that can be transferred from the BDD100K dataset to CARLA simulator without further fine-tuning, which proves the scalability of SSL to utilize large video datasets.



\small
\bibliography{bib/references}

\end{document}